%% file: main.tex
\documentclass[journal,11pt]{IEEEtran}

\input{packages}
\input{commands}

\begin{document}

\title{Use and Misuse of Machine Learning in Anthropology\thanks{{\bf Funding:} We would like to thank the National Science Foundation (NSF Grant DMS-1816917) for supporting this research. JC was partially supported by NSF Grant DMS-1944925, an Alfred P.~Sloan research fellowship and a McKnight Presidential Fellowship.}}

\author{
\IEEEauthorblockN{Jeff Calder\IEEEauthorrefmark{1}, Reed Coil\IEEEauthorrefmark{2}, Annie Melton\IEEEauthorrefmark{3}, Peter J.~Olver\IEEEauthorrefmark{1} \\ Gilbert Tostevin\IEEEauthorrefmark{3}, Katrina Yezzi-Woodley\IEEEauthorrefmark{3}}\\
\IEEEauthorblockA{\IEEEauthorrefmark{1}School of Mathematics, University of Minnesota, \{jwcalder,olver\}@umn.edu}\\
\IEEEauthorblockA{\IEEEauthorrefmark{2}Department of Sociology and Anthropology, Nazarbayev University, Kazakhstan, reed.coil@nu.edu.kz}\\
\IEEEauthorblockA{\IEEEauthorrefmark{3}Department of Anthropology, University of Minnesota, \{melto041,toste003,yezz0003\}@umn.edu}
}



\markboth{Use and Misuse of Machine Learning in Anthropology}%
{Shell \MakeLowercase{\textit{et al.}}: A Sample Article Using IEEEtran.cls for IEEE Journals}


\maketitle

\input{abstract}
\input{intro}
\input{ml_use}

\input{ml}
\input{ml_misuse}

\input{discussion}

\input{conclusion}

\bibliography{yezziwoodley.bib}
\bibliographystyle{IEEEtran}

\end{document}

%% file: packages.tex
\usepackage{amsmath,amsfonts}
\usepackage{algorithmic}
\usepackage{algorithm}
\usepackage{array}
\usepackage[caption=false,font=normalsize,labelfont=sf,textfont=sf]{subfig}
\usepackage{textcomp}
\usepackage{stfloats}
\usepackage{url}
\usepackage{verbatim}
\usepackage{graphicx}
\usepackage{cite}
\usepackage[dvipsnames]{xcolor}

\usepackage{hyperref}
\hypersetup{colorlinks,breaklinks,
             linkcolor=blue,urlcolor=blue,
             anchorcolor=blue,citecolor=blue}
\usepackage{xcolor}



%% file: commands.tex
\newcommand{\blue}{\color{blue}}

\def\blue#1{{\color{blue} #1}}

\graphicspath{ {images/} }

\renewcommand{\it}[1]{\textit{#1}} 

\usepackage{comment}

\usepackage{algorithm}
\usepackage{algorithmic}
\usepackage{dsfont}
\usepackage[bb=boondox]{mathalfa}

\newcommand{\x}{{\mathbf x}}

\newcommand{\y}{{\mathbf y}}
\renewcommand{\L}{{\mathcal L}}

\newcommand{\R}{{\mathbb R}}



%% file: abstract.tex
\begin{abstract}
Machine learning (ML), being now widely accessible to the research community at large, has fostered a proliferation of new and striking applications of these emergent mathematical techniques across a wide range of disciplines.  In this paper, we will focus on a particular case study: the field of paleoanthropology, which seeks to understand the evolution of the human species based on biological (e.g.~bones, genetics) and cultural (e.g.~stone tools) evidence. As we will show, the easy availability of ML algorithms and lack of expertise on their proper use among the anthropological research community has led to foundational misapplications that have appeared throughout the literature. The resulting unreliable results not only undermine efforts to legitimately incorporate ML into anthropological research, but produce potentially faulty understandings about our human evolutionary and behavioral past. 

The aim of this paper is to provide a brief introduction to some of the ways in which ML has been applied within paleoanthropology; we also include a survey of some basic ML algorithms for those who are not fully conversant with the field, which remains under active development. We discuss a series of missteps, errors, and violations of correct protocols of ML methods that appear disconcertingly often within the accumulating body of anthropological literature.  These mistakes include use of  outdated algorithms and practices; inappropriate testing/training splits,  sample composition, and textual explanations; as well as an absence of transparency due to the lack of data/code sharing, and the subsequent limitations imposed on independent replication. We assert that expanding samples, sharing data and code, re-evaluating approaches to peer review, and, most importantly, developing interdisciplinary teams that include experts in ML are all necessary for progress in future research incorporating ML within anthropology and beyond. 
\end{abstract}


%% file: intro.tex
\section{Introduction}

The purpose of anthropology is to better understand what it means to be human. This is an unimaginably broad field spanning all physical spaces that have been occupied by humans, from the present to the distant past. Though a plethora of frameworks are employed within anthropology, it is generally divided into four major subfields: biological anthropology, archaeology, socio-cultural anthropology, and linguistic anthropology. Biological anthropology broadly focuses on past, present, and future human biological variation, adaptation, and evolution. Archaeology studies human cultural evolution through the reconstruction of human behaviors based on the analysis of material culture remains. Socio-cultural anthropology examines the ways in which people navigate the world today. And, language as a cultural tool is the focal point of linguistic anthropology.

Given the breadth of the field of anthropology, this abbreviated survey will focus on how ML is currently impacting one particular subfield. Paleoanthropology is a multi-disciplinary field that brings together experts in Earth sciences, genetics, archaeology, biological anthropology and more to explore human evolution before the Holocene Epoch. The incorporation of ML into paleoanthropology follows a long tradition of adapting STEM methodologies to build inferences about the past (e.g., radiometric dating, ancient DNA sequencing, geometric morphometrics, etc.). 
Though lessons from this study may well impact other areas within anthropology where ML can be applied, we will restrict our attention to three areas of research within paleoanthropology that constitute the dominant sources of data in the field: the study of bone artifacts, stone artifacts, and the spatial associations between artifacts within and between sites. These are also the areas where we have both experience and expertise. We will focus on the analysis of bone modifications associated with butchering and the consumption of meat and marrow by early humans; behaviors related to the manufacture and use of stone (lithic) tools; and modeling the environments in which early humans lived.

Assemblages found at paleoanthropological sites are generally comprised of stones, bones, and other materials and are oftentimes fragmented. Deterioration over time of behaviorally-meaningful materials is an issue facing all archaeologists, let alone paleoanthropologists working in the deep past (more than 10,000 years ago). Due to this, paleoanthropologists seek to understand a complex past by extracting as much information from the remaining materials in the depositional record as possible, including quantifying and analyzing otherwise unremarkable fragments and pieces. Because paleoanthropology encompasses an array of disciplines and approaches too numerous to cover here, specifics on the nature of the data typical in the three aforementioned subfields will be explicated in Section II in the appropriate subsections.

The intended audience for this review paper consists of two almost disjoint groups of researchers: those versed in the basics of machine learning who are interested in new and promising directions of application, and those familiar with anthropology, in particular paleoanthropology, who are interested in the potential advances offered by modern machine learning tools.  As we will demonstrate by surveying the literature, while machine learning has begun to make inroads into paleoanthropology, its applications to date have often been compromised by failure to understand basic protocols and avoid common pitfalls. We argue that this underscores a need for interdisciplinary teams that combine researchers from both groups that can fully and correctly exploit the potentialities inherent in such an endeavor.  This is because machine learning experts are, by and large, not qualified to run a proper analysis of archaeological data, whereas 
anthropologists utilizing mathematical tools without full knowledge or informed input from experts can potentially lead to their misuse, thereby undermining their efforts to derive anthropologically meaningful outcomes. 
One platform for fostering such teams is the AMAAZE (Anthropological and Mathematical Analysis of Archaeological and Zooarchaeological Evidence) consortium (\href{https://amaaze.umn.edu/}{AMAAZE.umn.edu}), whose contributions to date are, in part, surveyed.  And since our intended readership is diverse, as we seek to engage as broad a readership as possible, we will present results and research that will, at times, be well known by one of the groups, but perhaps not by the other. We also envision that other social scientists, beyond anthropologists, may benefit from the lessons learned from this review.
 
For those who incompletely understand the mathematical foundations, ML carries a certain mystique, that is amplified by media reports of remarkable successes.  Even within the mathematically sophisticated research community, our lack of understanding of how ML algorithms work leads to a ``black box'' phenomenology where one judges the algorithms merely by some measure of success in assigned tasks. Often overlooked in the hype are the increasingly visible limitations of ML.  Even less commented on are the misuses of ML, in which basic procedures that are required to avoid  misleading and spurious classifications were not understood and/or followed. It is easy to achieve results that appear impressive to the ML novice if one does not follow the proper protocols and procedures.

The authors wish to emphasize that this cursory overview focuses on the missteps that have been made within anthropology and the applications of ML in terms of methods and data.  The ``appropriateness'' of the anthropological question and/or the archaeological method of investigation behind each study is beyond 
the scope of this paper. Specific points and potential pitfalls include proper use of training and testing data; the role and dangers of overfitting; the incorporation of bootstrapping; differences in machine learning algorithms; the concept of deep learning; the requirements underlying the specification of sample size, given that anthropology produces relatively small data sets; the influence of balance within the samples and how this and other considerations must be taken into account when interpreting results; and the necessary assumptions that must be met in order to apply machine learning methods in one's research.
Overall, four primary issues are observed when reviewing the existing ML studies in paleoanthropology: (1) train/test set contamination, (2) an absence, or incorrect application, of a train-test split, (3) lack of cross-validation and inappropriate measures of success, (4) a lack of transparency in the sharing of data and code that is standard practice among ML experts, and is essential for evaluating issues 1-3. In addition, many studies we reviewed contained inadequate or obfuscating textual explanations of ML methods, which made the evaluation challenging, especially when code and data were not shared.  

Nevertheless, despite the disconcerting findings concerning the current use of ML in the field, we are completely convinced that, when properly applied, machine learning promises a revolution in anthropology, particularly for our understanding of human evolution. The exceptional potential of machine learning for this field lies in overcoming as yet insurmountable classification problems that arise throughout anthropology, including those discussed here, identifying new feature sets conducive to machine learning algorithms, addressing the challenge of limited data sets through data augmentation and expansion into unsupervised learning, and by providing ways to expeditiously synthesize vast amounts and types of field data spanning large time scales or geographic regions in order to make sound inferences and interpretations.  
The goal of this paper is to help, in some small way, foster this revolution. Sections V and VI describe our proposals for how this can be effected.

%% file: ml_use.tex
\section{Overview of ML in paleoanthropology}

Many problems in anthropology and related fields involve the classification or categorization of objects to better understand how human behaviors and cultures changed across time and space. Paleoanthropologists have begun to use machine learning within their subfields to answer a variety of questions about human evolution. Below we summarize some of the current work in this field. 

\subsection{Bone modification studies}

\begin{figure*}[!t]
\centering
\subfloat[Fragmentation by Stone Tool]{\includegraphics[width=0.45\textwidth]{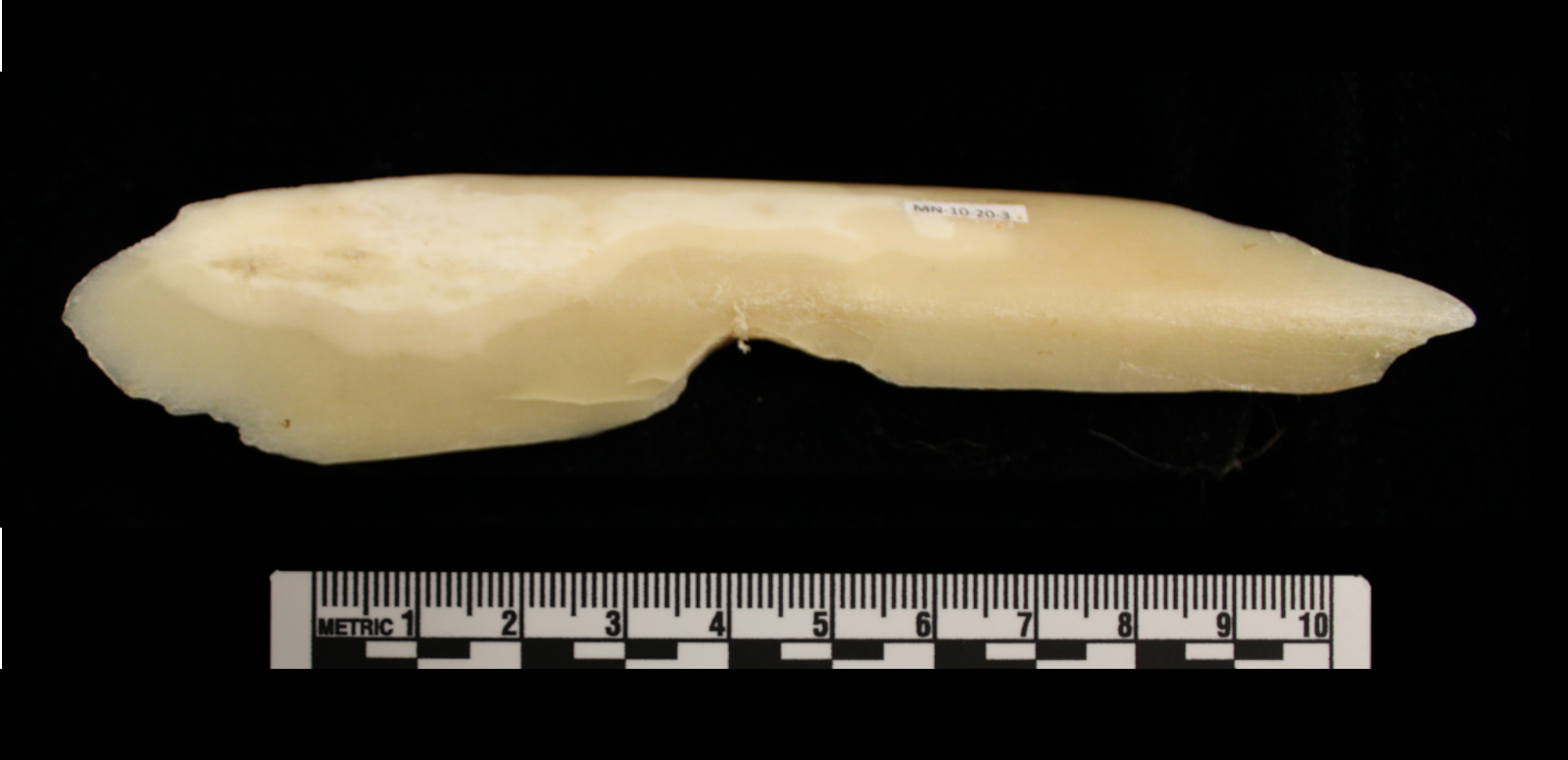}}
\hspace{9mm}
\subfloat[Fragmentation by Carnivore]{\includegraphics[width=0.45\textwidth]{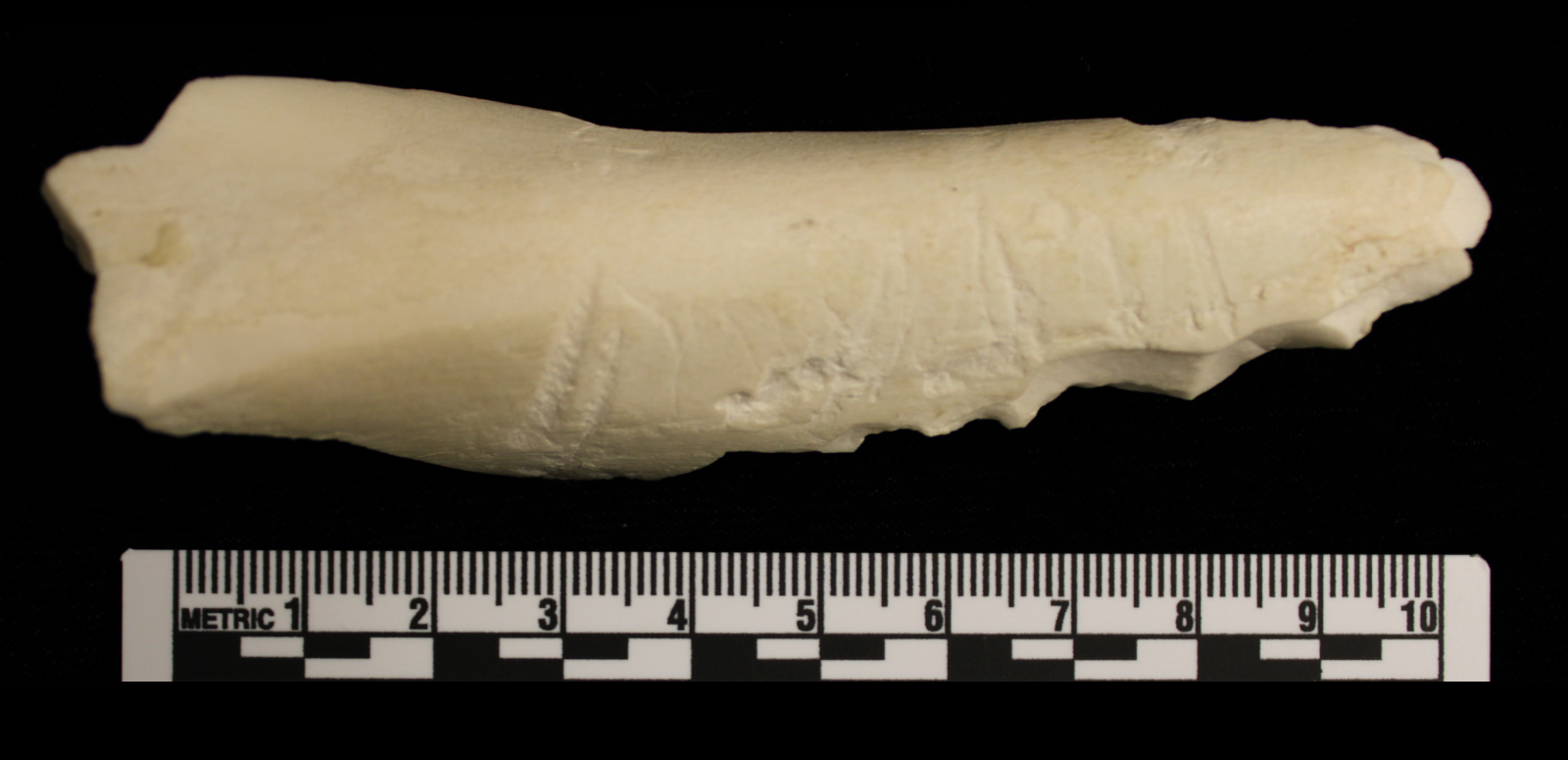}}
\hspace{9mm}
\caption{These fragments derive from elk limb bones that were experimentally broken. In image (a), the elk bone was broken by a human using stone tools. The indentation on the edge of the fragment indicates where the stone tool made impact with the bone. In the image b, the elk bone was fed to a spotted hyena at the Milwaukee County Zoo in Wisconsin. When carnivores chew on bones, their teeth create scores and pits on the surface of the bone and multiple adjacent indentations along the edge.}
\label{fig:agent}
\end{figure*}

Faunal remains are commonplace at paleoanthropological sites.  Fossil collections extracted from these sites are generally large by archaeological standards, and can contain over $10,000$ specimens which can provide a wealth of information for taphonomic analysis. Taphonomy is the study of what happens to an animal from the moment of death to the moment it is discovered by a paleontologist or anthropologist. This is often done through the examination of skeletal remains and the ways in which they have been modified through time. Analyzing bone fracture patterns and bone surface modifications (BSMs) is one way researchers reconstruct what happened in the past at these sites and ascertain early human subsistence patterns. Bone surfaces can be scratched, scraped, and otherwise damaged in a variety of ways. This can include stone tools that leave cut and percussion marks (Figure \ref{fig:agent}), carnivore mastication that leaves tooth scores and tooth pits (Figure \ref{fig:agent}), or the marks left behind by trampling bone in granular sediments. Bones can also be broken, for example, by humans or large carnivores that are interested in consuming embedded foods such as brains or bone marrow or by geological processes such as rockfall.

Since bones are one of the artifacts that occur in abundance at paleoanthropological sites, the identification of agents of bone breakage is essential to understanding how the site formed, how early humans evolved biologically and behaviorally, and how they interacted with their environment and with each other. However, long-standing debates over such identifications have yet to be resolved at important paleoanthropological sites such as are found in Dikika, Ethiopia (3.4 Ma) and Olduvai Gorge, Tanzania (1.8 Ma) \cite{yezzi2022using, dominguez2022case}.

Some researchers have applied machine learning to feature sets that are traditionally used in taphonomic analysis and are based on qualitative features as observed by the analyst and measurements taken manually \cite{dominguez2018distinguishing, dominguez2019successful, pizarro2020dynamic, moclan2019classifying, moclan2020identifying}. As examples, some of the data traditionally gathered by taphonomists include angles between features on the bones, dimensional measurements of bone surface modifications and bone fragments, and descriptive observations such as how straight or curved a linear BSM is or how jagged or smooth a fracture ridge is on a bone fragment (see Figure \ref{fig:series} for an illustration of such angle measurements). Most of these data are qualitative or measured using rudimentary tools such as calipers and handheld goniometers. 

Recent work has applied ML methods to the problem of classifying bone surface modifications according to the agents that produced them (e.g. humans, various carnivores, and trampling marks)  \cite{dominguez2018distinguishing, courtenay2019hybrid, dominguez2022case, abellan2021deep} (and other works by these authors), identifying human behavioral variation during butchering (e.g. using simple flakes with straight cutting edges versus retouched flakes that have a more serrated cutting edge) \cite{dominguez2018distinguishing}, differentiating marks made on fleshed and defleshed bones \cite{cifuentes2019deep}, exploring how captivity and domestication of dog species affects the morphology of the traces they leave behind \cite{yravedra2021use}, testing the efficacy of different methodologies \cite{courtenay2020obtaining}, and testing inter- and intra-observer variation during the process of feature extraction \cite{dominguez2019successful, courtenay2020obtaining}. Machine learning has also been applied to fracture patterns resulting from marrow extraction to identify whether carnivores or humans were responsible for breaking the bones \cite{moclan2019classifying, moclan2020identifying, yezzi2022using}.

Geometric morphometrics, which studies shapes through Cartesian landmarks, has recently been combined with machine learning to study 2D and 3D models of BSMs \cite{courtenay2019hybrid, courtenay2020obtaining, yravedra2021use} (and other works by these authors). And others have applied convolutional neural networks and transfer learning to images of BSMs \cite{dominguez2021use, dominguez2022case, abellan2021deep, cifuentes2019deep, pizarro2020dynamic} (and other works by these authors), including recent work using generative adversarial networks (GANs) for data augmentation \cite{dominguez2021use}.

Within the realm of fracture pattern analysis, machine learning has been applied to new feature sets resulting from recently developed methods for feature extraction from 3D models of bone fragments \cite{yezzi2021virtual, yezzi2022using}. These methods are highly accurate and replicable and can be applied to 3D models of any object. Thus, these methods can be used to address a wide array of anthropological questions and can be used by independent research teams for independent testing of anthropological applications. 

\subsection{Lithic technology}

\begin{figure*}[!t]
\centering
\subfloat[Textured 3D Models]{\includegraphics[width=0.25\textwidth]{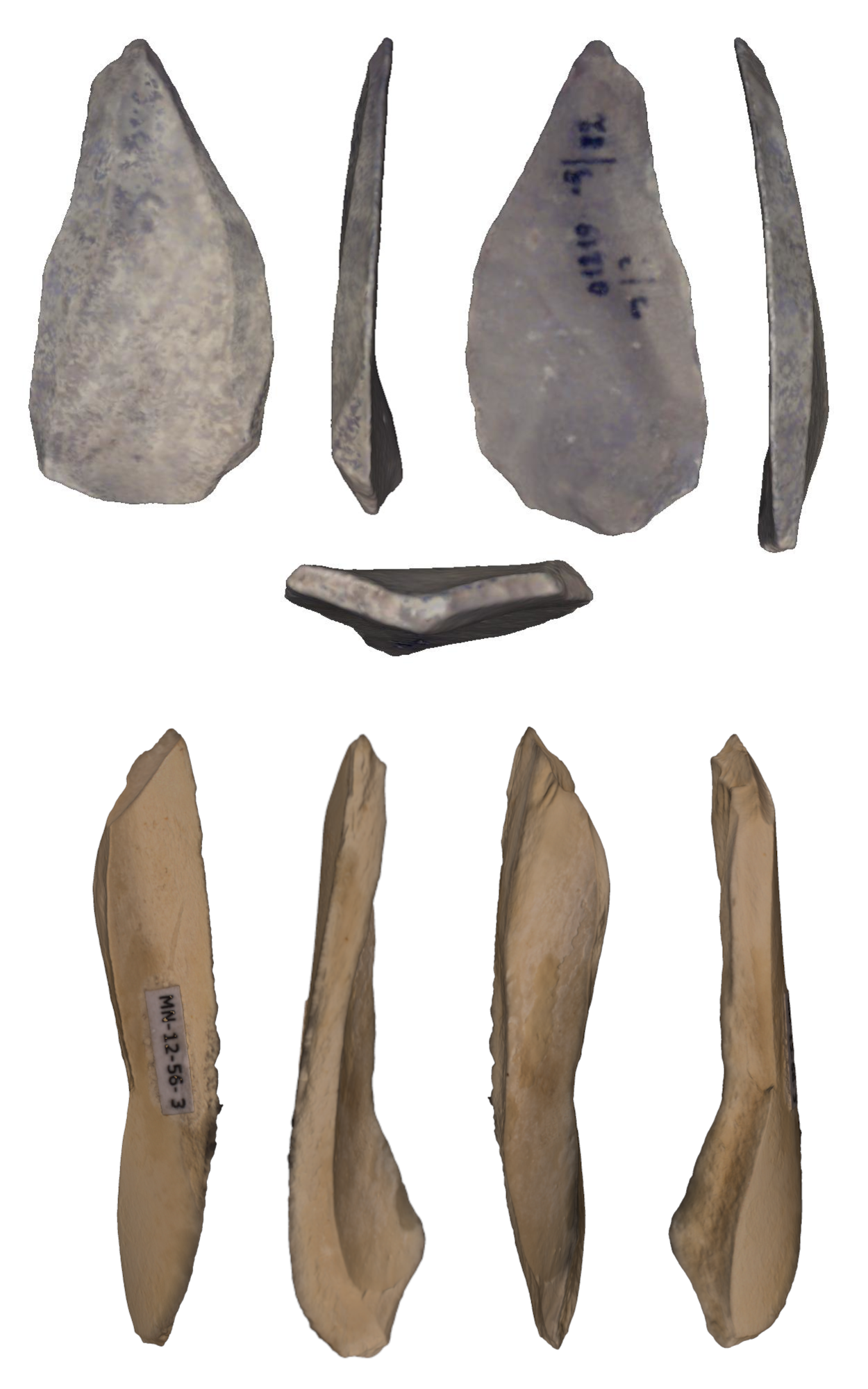}}
\hspace{9mm}
\subfloat[Segmented Models]{\includegraphics[width=0.25\textwidth]{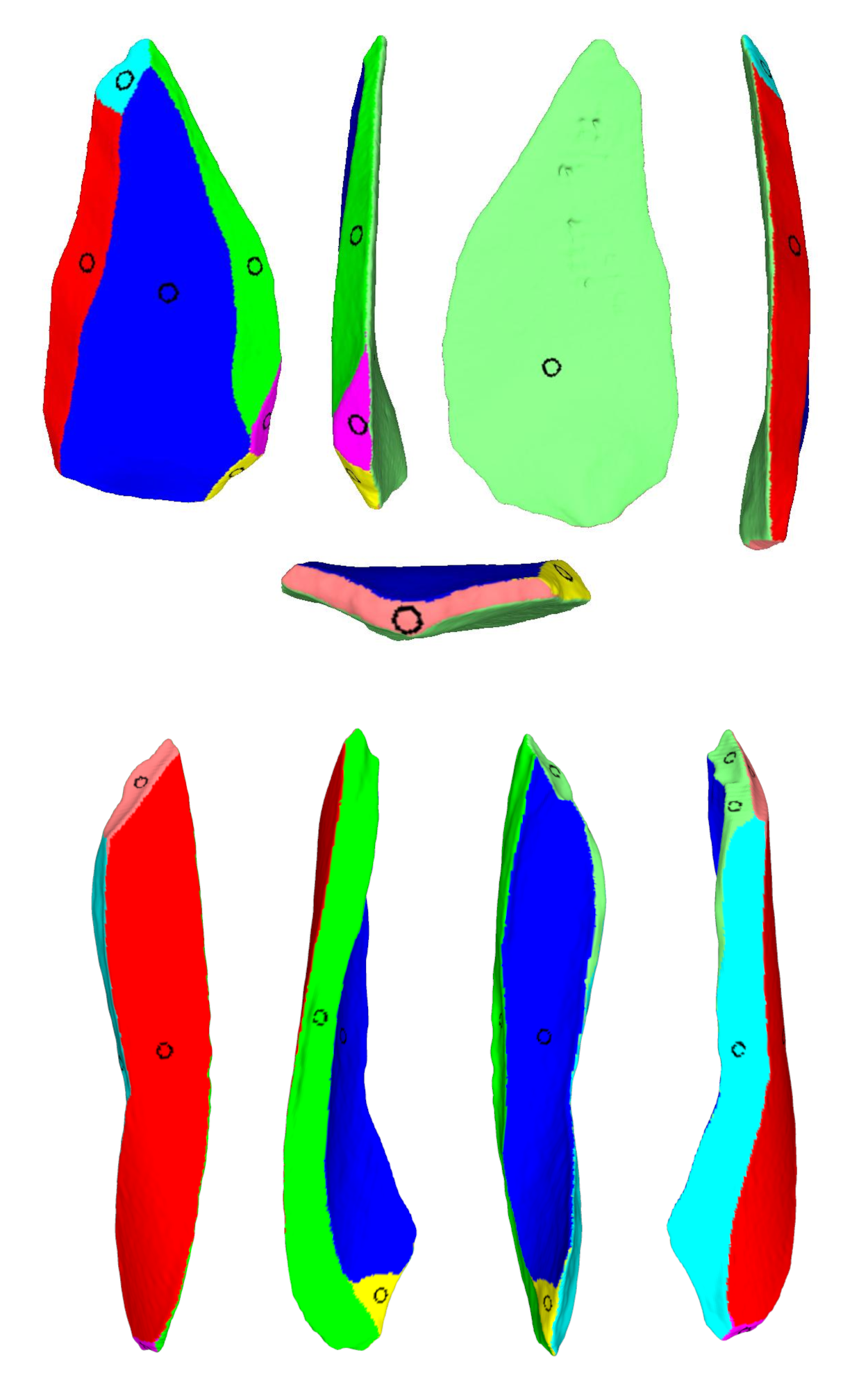}}
\hspace{9mm}
\subfloat[Angle Measurements Collected on 3D Models ]{\includegraphics[width=0.25\textwidth]{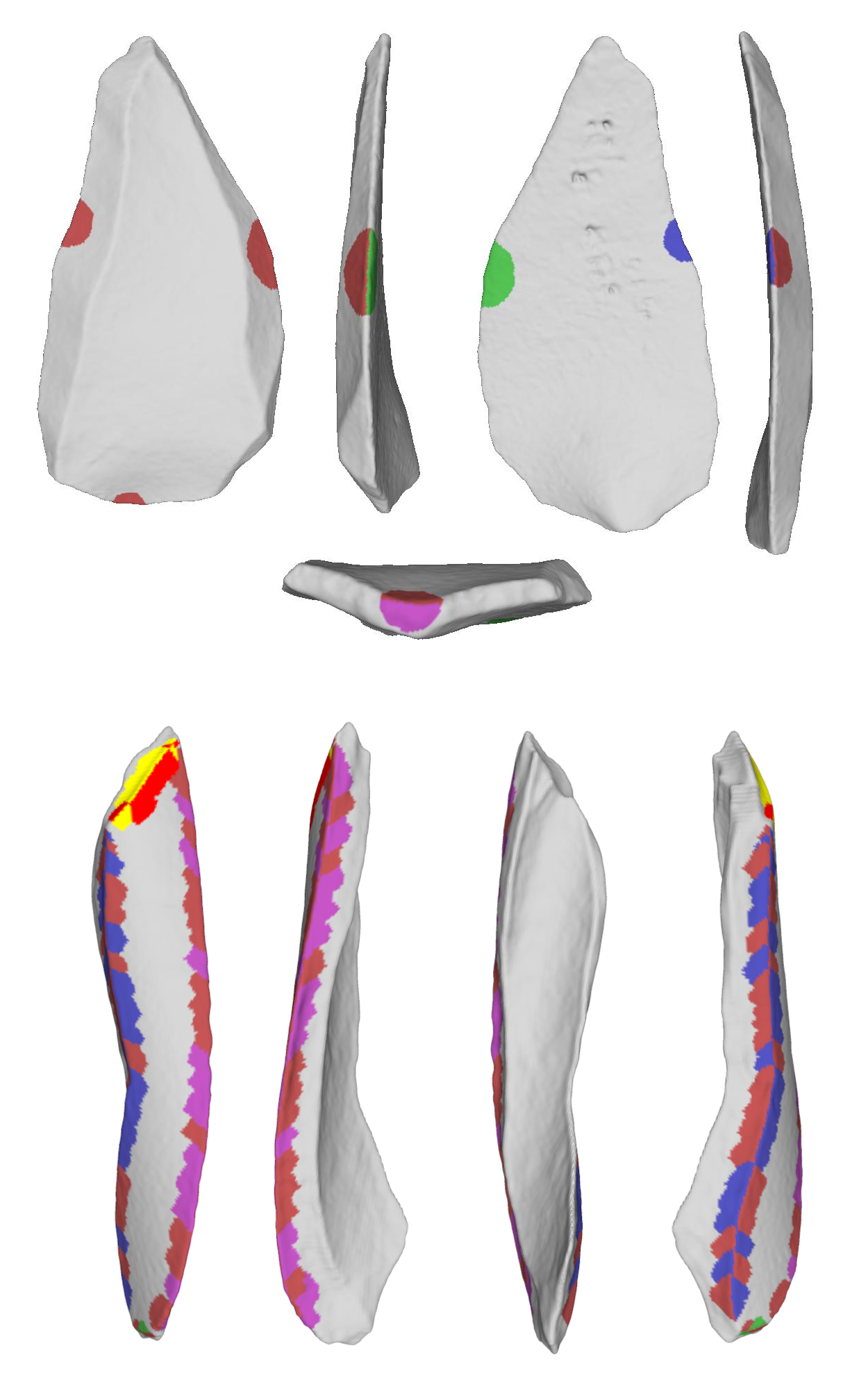}}
\caption{An illustration of the 3D models of archaeological materials and the types of data which can be collected from them: (a) textured 3D models of lithic (top) and bone (bottom) artifacts, (b) an illustration of how a researcher might separate out (i.e., segment) the differing planes which make up the geometry of these 3D objects, and (c) examples of angle measurements collected using the \emph{virtual goniometer} which can be taken on these types of objects. The pictured lithic artifact is from the site of  Stránská skála III (Czech Republic) and was 3D scanned by Gilbert Tostevin. A lithic artifact is made by striking a cryptocrystaline rock to create a Hertzian conchoidal fracture that separates a sharp flake tool from the parent core or nodule. The bone object, scanned by Katrina Yezzi-Woodley, is an experimentally-produced fragment from an elk produced by carnivore mastication.}
\label{fig:series}
\end{figure*}


Stone tool (lithic) technology is the most ubiquitous artifact type recovered from prehistoric archaeological sites.
As already mentioned, the archaeological record is largely fragmented (i.e., incomplete). How archaeologists study lithic artifacts is intended to compensate for the missing material. Metric and categorical data on key morphological attributes on lithic artifacts are collected in order to, for example,  pinpoint functional and cultural trends in tool production, reveal the production methods employed as sequential gestures of percussion, and characterize the transformation undergone by an artifact during its use, between initial creation and eventual discard.
As technology has progressed, so too have the methods lithic analysts use to study lithic artifacts (see Figure \ref{fig:series}). As one would expect, given the diverse anthropological questions one can hope to answer utilizing lithic materials (e.g., raw material sourcing, technological know-how and practice, population mobility) there are an even greater number of analytical approaches utilized today to answer those questions. While unable to detail the many developments in how to study lithic artifacts that have occurred in the last few decades, the following examples illustrate the varying types of data that can be collected to yield relevant behavioral information. Elemental analyses on lithics can reveal the sources of the raw materials used in tool production demonstrating mobility patterns exhibited by people in the past. The quantitative and qualitative analysis of artifactual debris of manufacture can be used to reconstruct the sequential steps in the reductive process of creating stone tools out of a raw nodule. In addition to a discontinuous view of the technological sequence, the continuous variables of shape and volume (among other things) can now be accurately analyzed due to the utilization of the latest 3D scanning technology allowing for more objective, quantitative assessments of lithic assemblages promoting comparability across research teams. ML is now one such ``new'' method being borrowed by archaeologists.

To date, ML has been applied in a number of lithic studies addressing a wide variety of anthropological questions: identifying heat-treated raw material nodules, a practice employed to improve the ease of working raw nodules into stone artifacts \cite{agam2021estimating}; identifying the materials worked by a stone tool according to the classification of the use-wear created on its edge  \cite{pedergnana2020evaluating,stevens2010practical}; predicting the original flake mass from variables on the striking platform in order to quantify the degree of resharpening (and thus the length of its use-life as a tool) \cite{bustos2021predicting}; predicting site formation conditions from the surface alteration of the site's lithic artifacts \cite{carranza2018study}; creating more quantitatively rigorous approaches to the creation of typologies for studying artifact shape through time and space \cite{nash2016use,macleod2018quantitative}; predicting the raw material of the stone tool from the cut marks produced by the edge \cite{cifuentes2021more}; identifying the geochemical signatures of geological sources of lithic raw materials as a means of studying prehistoric mobility and material selection criteria \cite{lopez2020projection,elliot2021evaluating}; distinguishing the flake products from different reduction strategies for exploiting the volume of a core \cite{gonzalez2020distinguishing}; distinguishing chronological manifestations of lithic behavior between the Middle and Late Stone Age in Africa through the presence vs.~absence of types within assemblages \cite{grove2020neural}; developing virtual knapping software \cite{orellana2021proof}; and quantifying lithic knapping skill acquisition for studying the evolution of human cognition \cite{pargeter2019understanding}.

\subsection{Environmental modeling}

Predictive models in the archaeological literature often explore wholly different questions, but most revolve around a similar theme: the reconstruction of past climates and environments and their effects on human evolution and behavior. 
ML has been applied to Geographic Information Systems (GIS) data to explore human-environmental interactions relating to niche construction, range expansion, biogeography, paleo-climate reconstruction, site use patterns, spatio-temporal analysis, interactions between early humans, or, often, combinations of many of these factors.  For example, ML has been employed to explore the socio-cultural and ecological factors relating to the geographic distributions of various techno-complexes (e.g. \cite{banks2008human}). Machine learning has also been applied to create a model of world population before the adoption of agriculture, which uses modern hunter-gatherers/foragers as analogs for past human groups \cite{hamilton2021reconstructing}.

Another primary application of ML on GIS data for archaeological purposes is remote sensing. Traditional remote sensing methods of satellite imagery, LiDAR, aerial photography, etc., involve time intensive exploratory data analysis and qualitative observations. Combining data gathered from archaeological survey, excavations, topographic maps, and geological sampling with more advanced satellite imagery (e.g., LandSat imagery) can revolutionize the search for prospective archaeological sites if the large sets of data can be synthesized and analyzed rapidly (e.g. \cite{coelho2021unsupervised, mertel2018spatial}).

\subsection{Machine learning in AMAAZE} 

In this section, we review recent applications of ML within our interdisciplinary consortium, the Anthropological and Mathematical Analysis of Archaeological and Zooarchaeological Evidence (AMAAZE)\footnote{See \blue{\url{https://amaaze.umn.edu}}}.  The motivating purpose of AMAAZE is to leverage advanced mathematical tools, particularly machine learning, time series analysis, and advanced geometric methods, to address fundamental questions in anthropology and human evolution that arise from the study of bones, lithics, and other artifacts, and thereby foster new and productive collaborations between mathematicians and anthropologists.  

As part of our research we seek to understand how the geometry of broken bone fragments is related to and helps distinguish among agents of breakage. 
Our studies are based on a large experimentally produced collection of bone fragments that are then used as proxies for what happened in the past. Our controlled samples are from elk, cow, sheep, and deer bones that have been broken by humans using stone tools, by spotted hyenas, and by simulated rockfall.  This collection has enabled us to begin to train, develop, and refine machine learning and geometry-based classification tools in preparation for the  analysis of field samples. 

The first step in this process was to develop new methods of feature extraction. In particular, the \emph{virtual goniometer} \cite{yezzi2021virtual} is a plug-in that can be used with the open access software Meshlab. This tool collects goniometric data with much greater accuracy and precision than the handheld pocket goniometer that is traditionally used for measuring angles on archaeological objects of interest. Data are automatically output to a $.csv$ file which prevents data recording errors and provides all the necessary information for replication by independent researchers ({\blue{see Fig. 2}}). 

In order to collect these data, we require 
3D models of bone fragments. To that end we developed the \emph{batch artifact scanning protocol} which creates triangulated surface meshes representing the physical objects rapidly ($<3$ minutes per fragment on average) \cite{yezzi2022batch}. The speed with which we are able to create 3D models can be attributed to the fact that we are able to simultaneously scan multiple fragments that are then automatically segmented and surfaced using Python scripts.

We have recently applied machine learning to a feature set extracted using the aforementioned tools \cite{yezzi2022using}. The purpose of this study was to differentiate bone fragments broken by hominins using hammerstone and anvil from those broken by spotted hyenas through mastication. The results are promising (average mean accuracy of $77\%$). This research is ongoing as we expand the experimental sample to include other agents of bone breakage and increase the size of the samples within each class. 

Once we have fine tuned and evaluated the ML algorithms for classification of our in-house collection of bone fragments, we will then apply our classifiers to samples and data gathered through field work at important paleoanthropological sites, namely Dmanisi, Georgia (1.8 Ma).

%% file: ml.tex
\section{Machine Learning}

We will next provide a brief overview, for non-experts, to some of the most popular machine learning models and algorithms, prior to our discussion on how machine learning technology is being used and misused in anthropology.

Machine learning (ML) is a type of artificial intelligence by which computers can develop the ability to perform tasks by learning from examples or experience, and are not \emph{a priori} coded with explicit instructions. ML methods learn from \emph{training data}, which includes \emph{features}  and \emph{labels}. The features can consist of both the actual data object, as well as data derived from the object, such as other information, measurements, and characteristics of the object. The labels associated with each data object are the targets for prediction or classification by the ML method. For example, in our work with bone fragments, the training data features contain various geometric measurements, most notably information about the break angles formed between the outside natural surface of the bone fragment and the broken surface, and surface curvature invariants, while the labels consist of the (known) actor of breakage. In general, the features can also include image data, such as computed tomography (CT) scans of the bone fragment, or a 3D model of the bone as a triangulated mesh.

ML methods can be broken down into three main categories: fully supervised, semi-supervised, and unsupervised learning. The distinction is based on how much labeled data is used by the algorithm. \emph{Fully supervised learning} algorithms learn from labeled data; that is, the algorithms use datasets that consist of both features and their corresponding labels to ``learn'' how to classify new datapoints. \emph{Unsupervised learning} refers to ML methods that use only features, and do not use any label information. Examples of unsupervised learning include clustering (i.e., grouping similar data points), data visualization, and dimension reduction. \emph{Semi-supervised learning} lies in between fully supervised and unsupervised learning; it makes use of both labeled and unlabeled data and is most useful when very little labeled data is available. We will focus our overview on fully supervised learning, as it is the most common in current paleoanthropological research. On the other hand, since the quantity of labeled data is limited in paleoanthropological collections, there is great potential for the use of unsupervised or semi-supervised techniques in future research. 

\subsection{Machine learning overview}
\label{sec:mlover}

In fully supervised ML classification, a \emph{training dataset} consisting of $n_{train}$ features $\x^{train}_1,\x^{train}_2,\dots,\x^{train}_{n_{train}}\in \R^D$ and corresponding labels $\y^{train}_1,\y^{train}_2,\dots,\y^{train}_{n_{train}}\in \R^C$ are used by the ML method to ``learn'' a classification rule that maps each $\x^{train}_i$ to its label $\y^{train}_i$. The classification rule is usually a parameterized function $f(\x;\omega)$, where $\omega\in \R^k$ are parameters that control the behaviour of $f$. The number of parameters $k$, which can be quite large, depends on the choice of ML model, and the choice of hyperparameters (see Section \ref{sec:hyper}) for that model. Training the ML method involves finding parameters $\omega$ so that $f(\x^{train}_i;\omega)$ is as close to $\y^{train}_i$ as possible, and this is often done by minimizing a \emph{training loss} of the form
\begin{equation}\label{eq:loss}
\L_{train}(\omega) = \frac{1}{n_{train}}\sum_{i=1}^{n_{train}} \ell(f(\x^{train}_i;\omega),\y^{train}_i),
\end{equation}
where $\ell(\y_1,\y_2)$ is a function that measures the discrepancy between the predicted and true labels. A simple example is $\ell(\y_1,\y_2) = \|\y_1-\y_2\|^2$, in which case the training loss is the mean squared error.  A variety of alternative loss functions can be utilized, depending on the application.  The optimal $\omega $ is commonly found or approximated using (stochastic) gradient descent.

The size of the training loss $\L_{train}$ is a measure of the success of the ML method in fitting the \emph{training data}, but in general does not give any indication of how the method will perform on new unseen data. The ultimate goal of ML is to ``learn'' classification rules that \emph{generalize well} to unseen data. In practice, the performance of the ML method on new data is measured by evaluating the model on a ``held-out'' \emph{testing dataset}, consisting of $n_{test}$ features $\x^{test}_1,\x^{test}_2,\dots,\x^{test}_{n_{test}}$ and labels $\y^{test}_1,\y^{test}_2,\dots,\y^{test}_{n_{test}}$ that 
have not been used in any way during training of the ML method.   The \emph{test loss} is given by
\[\L_{test}(\omega^*) = \frac{1}{n_{test}}\sum_{i=1}^{n_{test}} \ell(f(\x^{test}_i;\omega^*),\y^{test}_i),\]
where $\omega^*$ are the optimal weights chosen by the ML method during training. 

\begin{figure*}[!t]
\centering
\subfloat[Underfitting]{\includegraphics[width=0.25\textwidth]{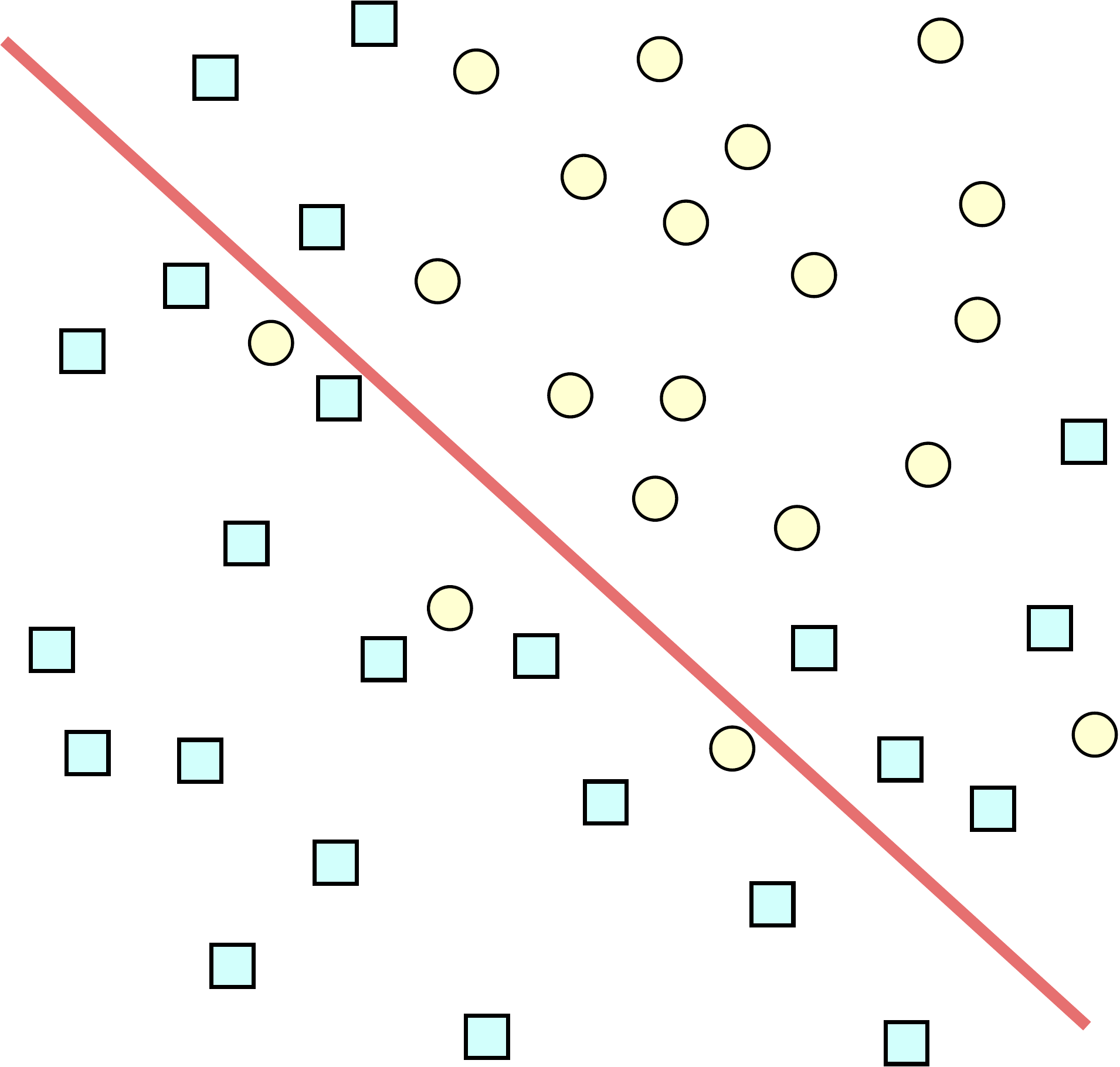}}
\hspace{9mm}
\subfloat[Appropriate fitting]{\includegraphics[width=0.25\textwidth]{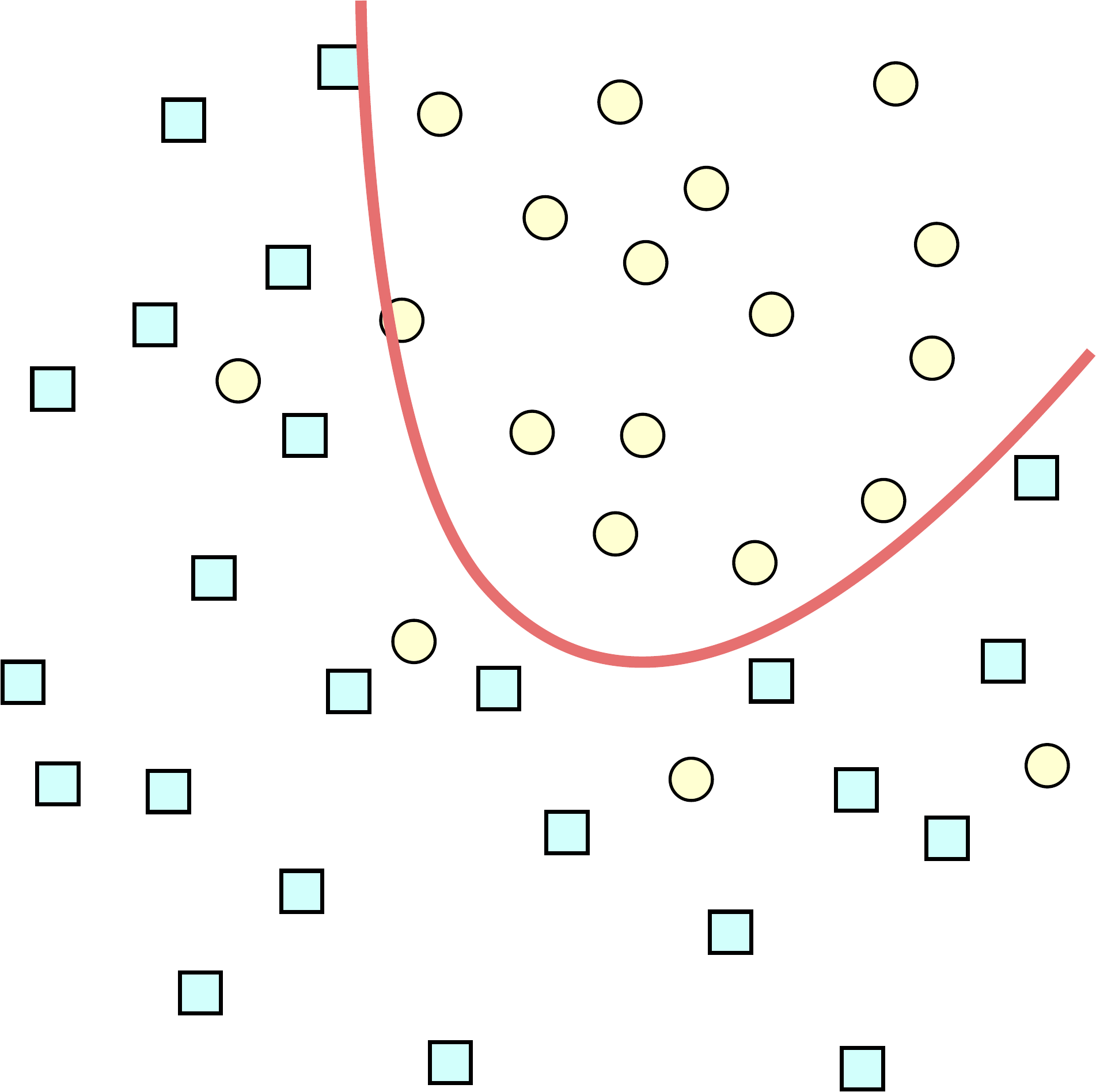}}
\hspace{9mm}
\subfloat[Overfitting]{\includegraphics[width=0.25\textwidth]{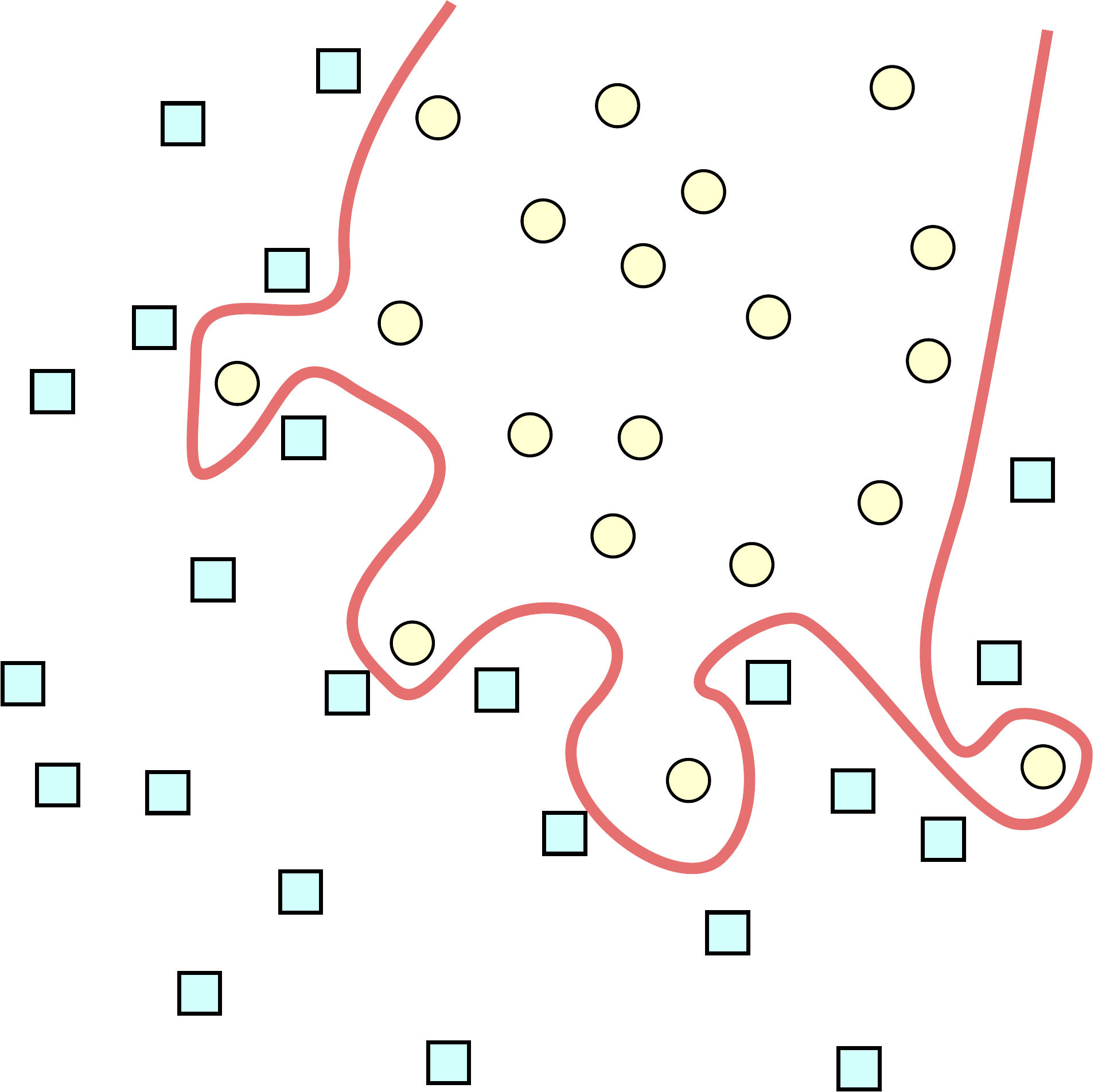}}
\caption{An illustration of (a) underfitting, (b) appropriate fitting, and (c) overfitting. In each case the red curve indicates the decision boundary learned by the ML method, and will be used to classify new unseen data points into one class or the other.}
\label{fig:fitting}
\end{figure*}

If the test loss $\L_{test}(\omega^*)$ is similar to the training loss $\L_{train}(\omega^*)$, then the ML method is said to \emph{generalize well}. In this case, if the loss is small, then the ML method is \emph{appropriately fitting} the training data, while if the loss is still large after training, then the method may be \emph{underfitting}. On the other hand, if the test loss is much larger than the training loss, then the model is \emph{overfitting} the training data and does not generalize well. Figure \ref{fig:fitting} serves to illustrate these three scenarios.  It is very important to emphasize that in a proper application of ML, the testing dataset must be independent of \emph{all} aspects of training, so that the test loss can be trusted as an unbiased estimation of model performance. 

In practice, the held-out testing dataset is obtained by making a \emph{train-test} split of the dataset \emph{before} training. One decides on the fraction of data to set aside for testing, say $25\%$, and then the dataset is split at random into a training set with $75\%$ of the data, and a testing set with $25\%$ of the data. It is essential to perform the train-test split prior to any steps used in training the ML method, and to use only the training data to train the model.  Some of the training data can be held out in a \emph{validation set}, in order to perform hyperparameter optimization or ensemble learning during training (see, e.g., Section \ref{sec:hyper}). 

In particular, it is important to ensure there is no contamination of data between the training and testing datasets. Indeed, many of the issues we discuss in the following sections stem from researchers either omitting the train-test split step, or mistakenly allowing data from the test set to contaminate the training set. There are subtleties in the train-test split that can lead to inadvertent contamination.  One must avoid training and testing data points with non-zero correlation between their labels, for instance, splitting on small scale features to classify large scale objects.  An example is the use of break data to classify bone fragments; see below for details. 

\subsection{Common ML methods}

An ML method is a particular choice of the parameterized function $f(\x;\omega)$ introduced in Section \ref{sec:mlover} (see Eq.~\eqref{eq:loss}), along with a training algorithm for determining the parameters $\omega$. There is now an ever-increasing plethora of ML methods, each of which has its advantages in certain applications. We describe some of the more commonly used and powerful methods below, and refer the reader to \cite{bishop2006pattern,lecun2015deep} for more details.  It is worth noting that the field of ML is rapidly evolving, whereby older algorithms are often replaced by improved contemporary methods; thus knowledge of the current literature is essential to well founded applications.  Furthermore, as new ML methods are developed, it would be worth revisiting earlier studies to see whether they can be replicated and, potentially, improved.  This is another reason for our emphasis on reproducibility and availability of data and code.

\subsubsection{\texorpdfstring{$k$}{k}-nearest neighbor classifier}

One of the most basic machine learning methods is the nearest neighbor classifier, in which a new data point with features $\x$ is classified by the label corresponding to the training datapoint whose feature vector $\x^{train}_i$ is most similar to $\x$. The similarity between feature vectors can be computed using various distance metrics on the features space $\R^D$, and common examples include any norm on Euclidean space, such as the Euclidean norm or $p$-norms (e.g.,  the $p=1$ Manhattan distance, or the $p=\infty$ Chebyshev distance), angular metrics like the cosine distance that use the dot product between feature vectors, and metrics for discrete feature vectors like the Hamming distance.  Often the nearest neighbor is not a reliable predictor of class membership, due to noise in the dataset, and so a superior classifier can be constructed by utilizing information from the $k$-nearest neighbors in feature space, where $k\geq 2$ is a ``hyperparameter'' specified in advance by the user. The labels of the $k$ neighbors are then combined (by majority vote or a weighted average) to perform the classification.

\subsubsection{Support vector machines (SVM)}

A support vector machine (SVM) uses a linear decision boundary to separate classes. In the case of two classes (i.e., binary classification), the SVM  classification rule is based on a real-valued linear function $f(\x;\omega)=\x\cdot \omega$ and a threshold $b \in \R$, and a data point $\x$ is in one class if $f(\x;\omega)>b$ and in the other if $f(\x;\omega)\leq b$. The parameters $\omega$ and $b$ are learned by maximizing the \emph{margin} of the linear classifier on the training dataset. Roughly speaking, the \emph{margin} measures how far the decision boundary is from the closest training datapoints (which are called \emph{support vectors}). Multi-class SVM with three or more classes works via the \emph{one-versus-rest} approach in ML, which is a general technique for constructing a multi-class classifier out of a binary one.  

Linear SVM works well only when the classes are \emph{linearly separable}, which means it is possible to find a line (or in higher dimensions, a hyperplane) so that the two classes (in binary classification) are on opposing sides of the line. The simple example in Figure \ref{fig:fitting} is not linearly separable. In order to handle such cases within the SVM framework, it is common to use the \emph{kernel trick}, whereby the training set features are augmented by additional nonlinear functions of the existing features, which lifts the data into a higher dimensional space, where linear SVM is applied. The goal is to choose a kernel for which the higher dimensional kernel features are linearly separable, even though they were not in the base space. Common kernels include polynomial functions, radial basis functions, and sigmoids.

\subsubsection{Decision trees and random forest}

Decision trees in machine learning use a binary tree decision making structure for classifying datapoints. Each node in the tree is a decision that performs a binary split on one feature in the dataset (i.e., is $x_1<1$?), and the classification of a new datapoint is determined by which leaf the data point arrives at after flowing through the decision tree. Decision trees are trained recursively in a greedy manner. At each step the method considers binary splits of all features, and selects the split that maximizes a measure of quality. The algorithm proceeds recursively until reaching a maximum tree depth. Decision trees have the advantage of mimicking some types of human decision making processes and their decisions can be more transparent and interpretable. However, they also risk overfitting, especially when they split on the same features repeatedly.  The random forest algorithm uses ensemble learning (see Section \ref{sec:ensemble}) to combine the outputs of multiple decision trees in a way that reduces overfitting and improves performance.

\subsubsection{Neural networks and deep learning}

Neural networks are ML models that loosely resemble the biological neural networks in human and animal brains. They are formed by arranging large numbers of individual neurons into interconnected layers. A single neuron is an affine function composed with a nonlinear activation function, i.e., $f(x;\omega,b) = \sigma(x\cdot \omega + b)$, where the weights $\omega$ and bias $b$ are tunable parameters. Common choices for the activation function include the rectified linear unit ReLU $\sigma(t)=\max\{t,0\}$ and the sigmoid $\sigma(t) = 1/(1+e^{-t})$. A single neuron can implement an SVM classifier, though its training does not seek the maximum margin classifier. Connecting multiple neurons together into layers can model more complicated nonlinear decision functions. \emph{Deep learning} simply refers to neural networks with at least 2 layers.

There are several kinds of neural networks, each designed for different types of data. A \emph{fully connected neural network} (or multi-layer perceptron, feed-forward neural network) can process any type of data that is represented as vectors in Euclidean space. They consist of several layers of neurons, connected so that each layers' outputs feed into the next layers' inputs. \emph{Convolutional neural networks} are specifically designed to process images. They are special cases of fully connected neural networks, where the affine functions in the neurons are replaced with the \emph{convolution} operation on 2D or 3D images, which is useful for extracting features therefrom. The convolution operation requires very few parameters, and explicitly encodes locality of image features, and translation equivariance, which increases the expressive power of convolutional neural networks for problems in computer vision. \emph{Recurrent neural networks} are specifically designed to process time series data (e.g., speech or handwriting recognition). The input to the network consists of the prior elements in the temporal data, and the output is a prediction that evolves in time. Finally, \emph{graph neural networks} refer to a wide class of neural networks that have been designed to process unstructured graph data, such as biological or social networks, or triangulated surfaces.

Deep learning methods are normally trained by minimizing a loss function like that in \eqref{eq:loss} with gradient descent. The \emph{backpropagation} algorithm, based on the chain rule for differentiation, is used to compute the gradients of the loss function in all the weight and bias parameters of the neural network. Since neural networks have the capacity to overfit, many types of regularization techniques have been proposed in the literature, including early stopping, dropout, batch normalization, and many others. With these modern techniques, deep learning generally does not overfit training data even when the neural networks are highly overparameterized, though the mathematical reasons for this are still poorly understood.



\subsection{Ensemble learning}
\label{sec:ensemble}

Ensemble learning is a general technique in ML for combining the results of multiple, possibly weak, ML classifiers together to obtain a stronger classifier with improved performance. There are many established techniques for ensemble learning, including bootstrap aggregation, boosting, stacking, and many others. In bootstrap aggregation, the weak classifiers are usually of the same type  (e.g., decision trees), but each classifier is trained on a different \emph{bootstrapped} version of the training set. The bootstrapped dataset is constructed by random sampling with replacement from the training set, and can also involve randomly sampling among the features of the data. The classifiers' performance is evaluated on the out-of-bag data (i.e., the training points not in the bootstrapped sample), and the multiple classifiers are combined based on their performance. One of the most widely used ensemble learning methods is the random forest algorithm, which combines the results of many random decision trees with bootstrap aggregation.

\emph{Boosting} refers to ensemble learning methods that operate incrementally, and focuses the training of future methods on the training points that were misclassified by previous models. One of the most widely used boosting methods is Adaboost\cite{freund1997decision}. \emph{Stacking} refers to combining multiple, possibly very different, ML methods by training another ML method to combine their predictions. 

We emphasize that ensemble learning must be performed on the training set alone. In particular, the evaluation of the multiple weak learners and learning how to combine them, must involve only the datapoints available in the training set. The testing set can only be used for a final evaluation of the ensemble learning method. 

\subsection{Cross-validation}

 The performance of ML methods can be dependent on the random selection of training and testing sets, especially when sample sizes are small. For proper evaluation of ML methods, it is important to use either $k$-fold cross validation, or to run the ML methods on many train-test splits chosen at random, in order to assess the variability in performance of the method with respect to changing the training and testing sets. $k$-fold cross validation splits the dataset at random into $k$ equal sized ``folds'', and trains the ML algorithm separately on each fold, taking the fold as the testing set, and the rest of the dataset as the training set.  
The average and standard deviation of accuracy scores over the $k$-folds or many random train-test splits should be reported.

\subsection{Hyperparameter optimization}
\label{sec:hyper}

Hyperparameters refer to parameters in ML methods that are used to control the learning process but are not optimized as a direct result of training the ML model. Examples of hyperparameters include the number of neighbors $k$  in the $k$-nearest neighbor algorithm, the choice of kernel in SVM, or the maximum tree depth in decision trees. In deep learning there are many hyperparameters, including the architecture of the neural network (i.e., number of layers, number of neurons per layer, etc.), the learning rate for gradient descent, how many training epochs to run, and the dropout rate, among many others. There are many hyperparameters in ensemble learning, controlling how the constituent ML methods are trained, evaluated, and combined. 

Optimizing hyperparameters can lead to improved results in machine learning. However, it is important that the process of optimizing hyperparameters does not utilize the testing accuracy in any way. Hyperparameter optimization must be performed using only the information present in the training set. Common techniques involve holding out part of the training set as a \emph{validation set} and using the validation accuracy to compare models trained with different hyperparameters. In a similar spirit, one can also use $k$-fold cross validation on the training set for hyperparameter tuning and model selection. 

%% file: ml_misuse.tex
\section{The misuse of ML in anthropology}

We have reviewed over 80 papers in the literature on applications of ML to anthropology and have identified a core set of fundamental mistakes that have been made. These mistakes render the results of many papers misleading and, in some cases, uninterpretable.   

\subsection{Train/test contamination}

One of the most common and serious types of mistakes we have observed are flaws in the ML workflow that lead to various amounts of contamination between the testing and training sets. The most egregious example of this is the use of bootstrapping to increase sample size \emph{before} the train-test split. Bootstrapping refers to sampling with replacement from a dataset and is a statistical technique with important applications in ensemble learning, where it is used to generate bootstrapped datasets to train constituent learners on. However, it should never be used to substantially increase the size of a dataset, nor should it ever be used on the whole dataset \emph{before} a train-test split (proper applications always involve bootstrapping the training set). Bootstrapping before a train-test split creates many duplicates of each datapoint, so that many (sometimes all) datapoints appear in \emph{both} the testing and training set. There is then no longer a held-out test set that can be used to evaluate the ML methods; consequently, reported accuracies should be interpreted as training accuracies, in which case high accuracies can merely indicate severe overfitting. In fact, it was shown in \cite{yezzi2022using} and \cite{mcpherron2021machine} that bootstrapping before the train-test split can produce accuracies close to $100\%$ on randomized datasets that contain no information.

There are several studies that have inappropriately used bootstrapping before train-test split, and report near perfect accuracy scores (e.g., \cite{dominguez2009new} reports $99.73\%-100\%$ accuracy). In \cite[p. 5]{dominguez2018distinguishing} it is stated, ``\textit{In order to provide the modelling with large training and testing/validation sets, the sample was bootstrapped 10,000 times, yielding a sample that is substantially bigger than BSM samples that one may encounter in archaeofaunal assemblages.}''.
In \cite[p. 2713]{dominguez2019successful} the author states, ``\textit{the sample was bootstrapped 1000 times to make it bigger and more similar to the samples that one may encounter in large archaeofuanal assemblages.}'' As \cite[p. 3]{mcpherron2021machine} point out in their critical response to \cite{dominguez2018distinguishing}, ``\textit{bootstrapping existing data cannot be used as a substitute for collecting more data''}. The reason to have large data sets when applying machine learning to classification problems is to capture the range of variability in each class. Resampling from existing data cannot accomplish this. 

As an explicit example, we note the dataset in \cite{dominguez2018distinguishing} contains $633$ BSMs. These were bootstrapped $10,000$ times to create a dataset with $10,000$ BSMs, though only $633$ are unique---the rest are duplicates. The $10,000$ sample dataset is then split into $70\%$ training and $30\%$ testing. 
For a given BSM, the probability that all of its bootstrapped copies end up in the same set (training or testing) is roughly $0.009$,\footnote{The probability is $\leq e^{-0.3\cdot 10,000/633}+e^{-0.7\cdot 10,000/633} \approx 0.009$.} and so the expected number of the 633 BSMs that are split properly into the train/test sets without contamination is less than $6$. Essentially all of the datapoints are in both the training and testing dataset, and so the results in \cite{dominguez2018distinguishing} are completely uninterpretable. 

Bootstrapping before the train-test split was used again by some of the same authors in \cite{moclan2019classifying}, where they inappropriately argued for its use in ML to increase the accuracy of ML classifiers. In addition, the authors of \cite{moclan2019classifying} make another serious mistake with their train-test split, which further contaminates the training set. The paper is concerned with classifying bone fragments by the agent of breakage, which can be hominin or animal in this case. They train ML classifiers to classify each break on a bone fragment, instead of classifying the entire fragment. Each fragment has several different breaks, and their train-test split is done on the break-level, meaning that each fragment can contribute breaks to both the training and testing set. This leads to a train/test contamination due to their use of fragment-level variables for classification, which are common to all breaks on a fragment. This issue was pointed out in \cite{yezzi2022using}, where it was shown that this type of break-level train-test split can also produce artificially high accuracies on randomized datasets that contain no information. These inappropriate uses of bootstrapping have diffused into the community, leading subsequent studies \cite{courtenay2019hybrid} to bootstrap their sample $1000$ times before the train-test split (they report near perfect accuracies around $99\%$).

A related issue involves applying data augmentation before the train-test split. Data augmentation in image classification involves increasing the size of the training set by applying random transformations to the training images (e.g., scaling, rotations, adding noise, color shifts, etc.), and is a very effective method for training deep neural networks to identify images in different situations. When applied properly data augmentation is done only to the training set and is usually done on each mini-batch during stochastic optimization. Several studies (e.g. \cite{abellan2021deep,dominguez2021use}) appear to be using data augmentation on the entire dataset to create a larger dataset \emph{prior} to the train-test split. This contaminates the training and testing data in a similar way to bootstrapping before a train-test split.

A related, though less serious, problem concerns applying certain types of preprocessing to the dataset before the train-test split. This can include standardizing the features (to be zero-mean with unit variance), applying principal component analysis (PCA) for dimension reduction, outlier detection and deletion, and so on. We want to stress that some of these missteps are relatively minor, and that data cleaning (i.e., removing erroneously recorded data, or datapoints with missing features) is a valid procedure to apply before the train-test split. Issues arise when the preprocessing goes beyond basic data cleaning. Standardizing the features, applying PCA, or certain types of outlier detection utilizes information from all the datapoints, including those that will later be assigned to the testing set, leading to train/test contamination. In a proper ML workflow, these preprocessing steps should be applied to the training set alone, and their maps can be recorded for use in testing.

The studies \cite{dominguez2019successful,dominguez2018distinguishing} apply standardization of the data before a train-test split, while the studies  \cite{courtenay2019hybrid, yravedra2021use} (and other works by these authors) apply PCA for dimension reduction before the train-test split. In \cite{elliot2021evaluating} the authors use the t-SNE embedding to visualize the dataset in two dimensions and manually remove ``outliers'' before the train-test split. Due to the difficulty interpreting the t-SNE embedding, the ``outliers'' removed could in fact be valid datapoints that are simply difficult to classify, thereby artificially increasing accuracy scores. In \cite{pedergnana2020evaluating} missing data was filled in with the median of the features from its class.

\subsection{Train-test split}

Another set of common mistakes concerns the train-test split. Many works either do not use a train-test split, in which case their accuracies should be interpreted as training accuracies, or they fail to ensure that the test set is not used in some way during training for tuning hyperparameters, model selection, or ensemble learning.

In their analysis of lithics, Grove and Blinkorn \cite{grove2020neural} use ensemble learning with an ensemble of $1000$ neural networks. However, they do not appear to retain a held-out testing set to evaluate the ensemble on. Each neural network is trained and evaluated on a random $85\%$/$15\%$ train-test split of the whole dataset, and they are combined based on their performance. Hence, every datapoint is used in training on average $850$ of the neural networks, and so there is no held-out testing data to evaluate the model. Nash and Prewitt \cite{nash2016use} do not use a train-test split; they train their models on the whole dataset. As such, the tables in their paper are misleadingly reporting training accuracy. They do test their models on a new testing set of $5$ datapoints, but this is too small for proper evaluation. MacLeod \cite{macleod2018quantitative} does not  perform a train-test split, as the entire data set is included within the confusion matrix. His accuracy results of close to $100\%$ are thus training accuracy and indicative of overfitting.  

A related issue has to do with model selection. In \cite{dominguez2019successful} the author advocates for using as many ML methods as possible for any problem at hand, and to then choose the best one. This approach aligns well with the spirit of ensemble learning, however, the authors do not employ ensemble learning, and they instead evaluate their plethora of models on the testing set. It is important to use a validation set for model selection in the ensemble learning framework, so that there is a held-out test set for evaluating the ``best'' model. Similar issues appear in \cite{dominguez2022case}, which uses two layers of ensemble learning, whereby they correctly train 18 ensembles that each reach about 95\% accuracy, and then hand pick 4 of the ensembles that performed best to create a super-ensemble that predicts by majority vote and obtains 100\% accuracy. Similarly, in \cite{abellan2021deep} the authors experiment with a large number of models and select the best ones, based on testing accuracy, for further supervised learning experiments. 

\subsection{Cross-validation and measurements of success}

Several studies do not use cross-validation or multiple train-test splits. In the context of small sample sizes, the testing accuracy from one split may be a poor indicator of expected model performance. For example, in \cite{mertel2018spatial} the authors consider only one train-test split on a dataset with $338$ datapoints. 

We have also observed studies drawing incorrect or unsupported conclusions from testing models on new data. In \cite{dominguez2019successful}, the author tests their model's predictions against those of three expert analysts in the field, and finds that their model performed vastly differently than the experts, agreeing very well with one expert and very poorly with another. They concluded that this provides evidence of high inter-observer error. However, an equally valid explanation is that their model is overfitting the training data and does not generalize well. The study, as published, therefore cannot distinguish between inter-observer error and model overfitting.

%% file: discussion.tex
\section{Discussion}

We have discovered a large number of cases appearing in the published paleoanthropological literature in which machine learning methods were misused, leading to faulty if not wrong conclusions and misleading estimations of success.  In order to avoid further invalid applications of ML, we advocate for procedures that include interdisciplinary collaboration, well founded and reliable peer-review, archiving of data and code in readily available repositories that can be used for replication and further analysis by independent research teams, and increasing sample sizes. 

\subsection{Inter-disciplinary teams}

The proper application of ML methods in research requires a wide range of expertise, including familiarity with the mathematical foundations of the subject. The ease at which modern software packages can be used to implement ML methods should not be regarded as a replacement for domain-area expertise. A number of the studies we reviewed relied on out-dated or inadequate algorithms that are no longer employed in modern ML practice. This does not mean the methods are incorrect; they are simply hard to evaluate and the results could be substantially improved by inter-disciplinary collaboration. Archaeologists are typically not trained in computer science, nor mathematics, and certainly do not have the depth-of-knowledge acquired by experts in machine learning.
Such research is, at its foundation, inter-disciplinary in nature, and thus is best conducted with inter-disciplinary teams of researchers, where ML experts can propose and vet the appropriate methods and protocols, and thereby identify and avoid common mistakes. Otherwise, the great potential that ML may yield in understanding human behavior in the past will never be realized (see \cite{huggett2021algorithmic} for further discussion). 

\subsection{Peer review process}

The frequency of common ML mistakes in the papers we reviewed suggests to the authors that there is an absence of a rigorous and informed peer review structure in place within existing archaeological journals that can appropriately vet the ML protocols employed. This situation is particularly evident when the peer review process allowed the publication of inaccurate, inappropriate, or obfuscated ML methods that would not have been acceptable in journals regularly utilized by ML experts. 

We would advocate for archaeological journals that frequently publish ML studies to include one or more ML experts on their editorial board and for ML studies to be peer reviewed by experts in ML as well as archaeology. Without ML expertise at the highest levels in the journal hierarchy, poor quality ML studies will continue to be published in high quality archaeology journals, damaging the field and the journals' reputations. We also advocate for the creation of new cross-disciplinary journals focused on applications of ML in archaeology and other areas where ML is being actively used to address these concerns.

\subsection{Reproducibility}

A majority of the papers we reviewed suffered from issues of reproducibility. Many papers (e.g., \cite{lopez2020projection,stevens2010practical}) had inadequate or confusing textual explanations of the ML methods, making comprehension of the appropriateness of the ML application difficult. Access to raw data and code enables the reader to clarify such confusions. Unfortunately complete data and code were frequently not available, so we were not always able to verify with certainty the exact ML methods that were used and the mistakes that were made. Of the papers cited in this review, the citations that share no, or incomplete data, and/or no code include \cite{banks2008human, bustos2021predicting, carranza2018study, courtenay2019hybrid, courtenay2020obtaining, dominguez2018distinguishing, dominguez2019successful, dominguez2022case, gonzalez2020distinguishing, hamilton2021reconstructing, kondo2012developing, lopez2020projection, moclan2020identifying, nash2016use, pizarro2020dynamic, stevens2010practical, yravedra2021use, abellan2021deep, cifuentes2021more, moclan2019classifying}. 

Not only does the sharing of data and code aid the reader in understanding the published works, but it also enables independent replication which is a hallmark of the scientific process and fundamental for vetting and advancing research. We feel that sharing code and data needs to become a standard practice within anthropology for those who apply ML in their anthropological research (or even writ large).

\subsection{Sample sizes}

Another complicating factor in any archaeological study but particularly those in paleoanthropological contexts is sample size. ML methods that rely on large sample sizes may not perform well on the smaller data sets common in the field. With small sample sizes, bias can creep into trained models, and even models with seemingly good testing accuracy may not generalize to field data.  Archaeological studies are limited to the archaeological record (i.e., what is preserved and recovered), so there are frequently strict limitations on sample sizes (usually in the hundreds or at most thousands of datapoints). This does not mean ML cannot be applied, but researchers should be aware of sample size issues and be transparent about how far their conclusions actually reach. To harness the power of modern machine learning techniques, which thrive on vast amounts of training data, researchers should be working to increase sample sizes in the future.

%% file: conclusion.tex
Overall, while daunting given the complexities of its usage and the underlying mathematics, ML has much to offer paleoanthropological studies. This is particularly evident in topics such as bone taphonomy which have been hotly debated for decades despite the application of non-ML statistical techniques. Again, we wish to emphasize that this review is not meant to discourage researchers from using ML. Instead, we are detailing the aforementioned warnings and recommendations in order to ensure that this burgeoning growth in ML applications continues appropriately. As with any new analytical technique development (or borrowing), there is a ``learning" period during which researchers must figure out what is appropriate or not. Paleoanthropologists have much to learn in terms of how to appropriately apply ML to their studies, but collaborations with ML experts will greatly expedite this learning period. 
By working together, we can build more reliable test sets, identify the types of data and resources that need to be shared in order to reproduce studies, define the best ways in which to share those data, and design approaches to effectively evaluate research outcomes.  

Our individual paleoanthropological datasets may be limited, but practicing data-sharing and open code access through platforms such as Github  will only improve future studies (e.g., larger, aggregated datasets; improved ML algorithms). 
The authors are excited to see the growth of interdisciplinary research and research teams which will result from the growing applications of ML in paleoanthropology and archaeology. With expanded research networks (and the accompanying new perspectives), we expect to see many new, stimulating questions asked and answered.